\documentclass[conference]{IEEEtran}
\IEEEoverridecommandlockouts
\usepackage{cite}
\usepackage{amsmath,amssymb,amsfonts}
\usepackage{algorithmic}
\usepackage{graphicx}
\usepackage{textcomp}
\usepackage{xcolor}
\usepackage{multirow}
\usepackage{tipa}
\def\BibTeX{{\rm B\kern-.05em{\sc i\kern-.025em b}\kern-.08em
    T\kern-.1667em\lower.7ex\hbox{E}\kern-.125emX}}
\begin{document}

\title{Classifying Graphemes in English Words Through the Application of a Fuzzy Inference System}

\author{\IEEEauthorblockN{Samuel Rose}
\IEEEauthorblockA{\textit{School of Computer Science} \\
\textit{University of Hull}\\
Hull, UK \\
s.p.rose-2017@hull.ac.uk}
\and
\IEEEauthorblockN{Chandrasekhar Kambhampati}
\IEEEauthorblockA{\textit{School of Computer Science} \\
\textit{University of Hull}\\
Hull, UK \\
c.kambhampati@hull.ac.uk}
}

\maketitle

\begin{abstract}

In Linguistics, a grapheme is a written unit of a writing system corresponding to a phonological sound. In Natural Language Processing tasks, written language is analysed through two 
different mediums, word analysis, and character analysis. This paper focuses on a third approach, the analysis of graphemes. Graphemes have advantages over word and character 
analysis by being self-contained representations of phonetic sounds. Due to the nature of splitting a word into graphemes being based on complex, non-binary rules, the application of fuzzy logic \cite{b22} would provide
a suitable medium upon which to predict the number of graphemes in a word. This paper proposes the application of a Fuzzy Inference System to split words into their graphemes. This Fuzzy
Inference System results in a correct prediction of the number of graphemes in a word 50.18\% of the time, with 93.51\% being within a margin of +- 1 from the correct classification. Given the variety in language, graphemes are tied with pronunciation and therefore can change depending on a regional accent/dialect, the +- 1 accuracy represents the impreciseness of grapheme classification when regional variances are accounted for. To give a baseline of comparison, a second method involving a recursive IPA mapping exercise using a pronunciation dictionary was developed to allow for comparisons to be made. 
\end{abstract}

\section{Introduction}
A grapheme is the smallest functional unit of a writing system \cite{b6}. In English, graphemes are the written representation of a
phonetic sound, phonemes. In English, a grapheme varies in length from 1-4 characters. These work as the building blocks for words in English, 
giving both structure for how a word is spelt, and representing a words pronunciation. The decoding of words into graphemes is a fundamental process in learning English and is included in the National Curriculum for Key Stage 1 Students in England \cite{b7}. 

While the English language contains 44 phonemes \cite{b10}, there is no fixed number of graphemes due to the mixed linguistic base of English words. For this project,
 a figure of 284 graphemes is used. This includes 89 graphemes in the ``main system'' which account for the majority of words, and a further 195 graphemes that cover niche cases not covered in the ``main system'' \cite{b4}. 

\subsection{Grapheme Decoding Challenges}
\label{Challenges}
One of the challenges in decoding graphemes in English is, due to the vast number of graphemes compared to phonemes, multiple graphemes represent each
 phoneme and the same grapheme can represent multiple phonemic sounds depending on the context. One example of this is the ⟨ei⟩ grapheme which represents three 
 different phonemes, /i:/ where it is sounded as a long e as in ``deceive'', /\textipa{eI}/ where it is sounded ay as in ``weigh'', and also /\begin{math}\mathcal{E}\end{math}/ where it 
 is sounded as a short e as in ``heifer''. Another challenge in decoding a word's graphemes is that characters in some graphemes are contained as part of another grapheme. 
 Using the ⟨ei⟩ grapheme as an example again, ⟨ei⟩ is a grapheme consisting of the characters e and i, ⟨e⟩ and ⟨i⟩ are also single-character graphemes, and the characters ei are part of 
 the ⟨eir⟩ and ⟨eigh⟩ graphemes. 

Because of these challenges in decoding graphemes, the decoding of a word's graphemes does not fit into discrete categories. Due to this, we hypothesise that by applying Fuzzy Set Theory,
English words can be computationally decoded into their graphemes by applying fuzzy sets and fuzzy inference systems. 

\section{Development of Fuzzy Sets}
\subsection{Grapheme Count Probabilities}
The first step in developing fuzzy sets for deciding a word's graphemes is to work out how many graphemes are in a word. To achieve this task, frequency analysis can be undertaken on an
annotated corpora of English words with graphemes. For this task the training corpus used is the Google 10,000 \cite{b21}, a corpora of the top 10,000 words ever searched in Google, with the EnglishGrammarHere10000 as a validation corpus \cite{b11}. Each word in this list
was then decoded manually using the rules set out for decoding graphemes in the National Curriculum \cite{b7}. The decision to choose the top 10,000 most common English words is that 
according to analysis undertaken by the Oxford English Corpus \cite{b18}, the 7000 most common words in English account for approximately 90\% of word usage, making a corpus of 10,000
a suitable size for computational analysis. 

\begin{figure}[htbp]
\centerline{\includegraphics[width=8cm]{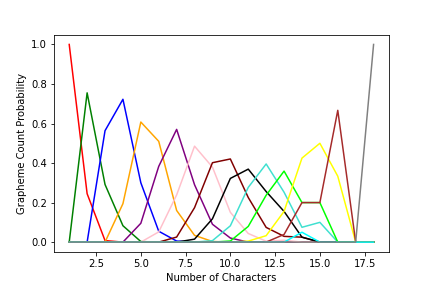}}
\caption{Probabilities of Grapheme count compared to word length}
\label{Grapheme Probabilities}
\end{figure}

To obtain frequency distributions for the number of graphemes in a word, the corpus was split into subsets of words length 1-$n$ where $n$ is the longest word in the corpus. In the corpus,
the longest word was the word ``telecommunications'' at 18 characters long. In each subset, the number of graphemes for each word, from 1 to $n$, is counted. After this process was completed,
the relative probabilities were calculated. This was calculated by dividing the word by the number of words of length $n$ which are in the dataset. These probabilities were then plotted, showing
that the number of characters in each grapheme roughly follows a normal distribution, shown in Figure \ref{Grapheme Probabilities}. Based on this result, the conclusion made is that pursuing this line 
of experimentation has significant potential for yielding a positive result. 

\subsection{Development of Fuzzy Sets}
With the number of graphemes when compared to word length loosely following a normal distribution, by calculating the mean,  $\mu$, and standard deviation, $\sigma$, for the number of 
characters, vowels and consonants for words of $x$ Graphemes, these can be seen, rounded to 3 decimal places, in Table \ref{MeanStDevTable}.  Having calculated the mean and standard
deviation for words consisting of 1-14 graphemes, the first 12 of these have a standard deviation greater than zero and can be approximated using a normal distribution. The formula of a
normal probability density distribution, $f(x)$ is as follows: 

\begin{equation}
f(x) = \frac{1}{\sigma\sqrt{2\pi}}e^{-\frac{1}{2}(\frac{x-\mu}{\sigma})^2} 
\end{equation}

\begin{table*}[]
\caption{Grapheme Means and Standard Deviations}
\centering
\begin{tabular}{|l|l|l|l|l|l|l|}
\hline
\multicolumn{1}{|c|}{\multirow{2}{*}{\begin{tabular}[c]{@{}c@{}}Grapheme\\ Count\end{tabular}}} & \multicolumn{2}{c|}{Characters} & \multicolumn{2}{c|}{Vowels} & \multicolumn{2}{c|}{Consonants} \\ \cline{2-7} 
\multicolumn{1}{|c|}{} & \multicolumn{1}{c|}{Mean} & \multicolumn{1}{c|}{\begin{tabular}[c]{@{}c@{}}Standard \\ Deviation\end{tabular}} & \multicolumn{1}{c|}{Mean} & \multicolumn{1}{c|}{\begin{tabular}[c]{@{}c@{}}Standard \\ Deviation\end{tabular}} & \multicolumn{1}{c|}{Mean} & \multicolumn{1}{c|}{\begin{tabular}[c]{@{}c@{}}Standard \\ Deviation\end{tabular}} \\ \hline
1 & 2 & 0.508 & 1.219 & 0.608 & 0.781 & 0.608 \\ \hline
2 & 3.062 & 0.680 & 1.245 & 0.593 & 1.817 & 0.671 \\ \hline
3 & 4.202 & 0.820 & 1.573 & 0.651 & 2.629 & 0.720 \\ \hline
4 & 5.556 & 0.897 & 2.031 & 0.730 & 3.525 & 0.758 \\ \hline
5 & 6.853 & 0.966 & 2.550 & 0.752 & 4.303 & 0.848 \\ \hline
6 & 8.063 & 1.036 & 3.058 & 0.810 & 5.005 & 0.913 \\ \hline
7 & 9.221 & 1.072 & 3.537 & 0.891 & 5.684 & 0.962 \\ \hline
8 & 10.290& 1.099 & 4.020 & 0.869 & 6.270 & 1.014 \\ \hline
9 & 11.298 & 1.011 & 4.560 & 0.892 & 6.738 & 0.989 \\ \hline
10 & 12.18 & 0.967 & 4.924 & 0.882 & 7.260 & 0.925 \\ \hline
11 & 13.396 & 1.067 & 5.438 & 0.873 & 7.958 & 0.898 \\ \hline
12 & 14.125 & 0.957 & 5.625 & 0.806 & 8.5 & 0.632 \\ \hline
13 & 14 & 0 & 6 & 0 & 8 & 0 \\ \hline
14 & 18 & \begin{tabular}[c]{@{}l@{}}Div/0\\ Error\end{tabular} & 8 & \begin{tabular}[c]{@{}l@{}}Div/0\\ Error\end{tabular} & 10 & \begin{tabular}[c]{@{}l@{}}Div/0\\ Error\end{tabular} \\ \hline
\end{tabular}
\label{MeanStDevTable}
\end{table*}

\begin{figure}[htbp]
\centerline{\includegraphics[width=8cm]{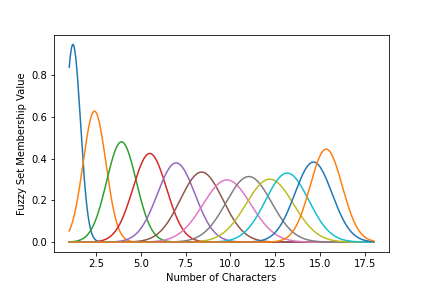}}
\caption{Frequency Sets for Grapheme count compared to word length}
\label{FrequencySets}
\end{figure}

Using the Mean and Standard deviations calculated in Table \ref{MeanStDevTable}, frequency sets for words with 1-12 graphemes were plotted, shown in Figure \ref{FrequencySets}. 

\subsection{Fuzzy Inference System}
Having plotted the frequency sets for words with 1-12 graphemes, a Mamdani Fuzzy Inference System was developed \cite{b13}. This system is designed to have three inputs, word length, vowel count, 
and consonant count. These are represented by membership functions with the mean and standard deviations outlined above. Using a set of fuzzy rules
a mapping is developed that links the input functions to output functions. The result of this is an aggregate set correlating how much the inputs relate to each output rule. This set is then
evaluated to result in a non-fuzzy output, in this case, the output is a number, with a precision of 2 decimal points in the range [1-14]. 

\subsubsection{Input Membership Functions}
\begin{figure}[htbp]
\centerline{\includegraphics[width=6.5cm]{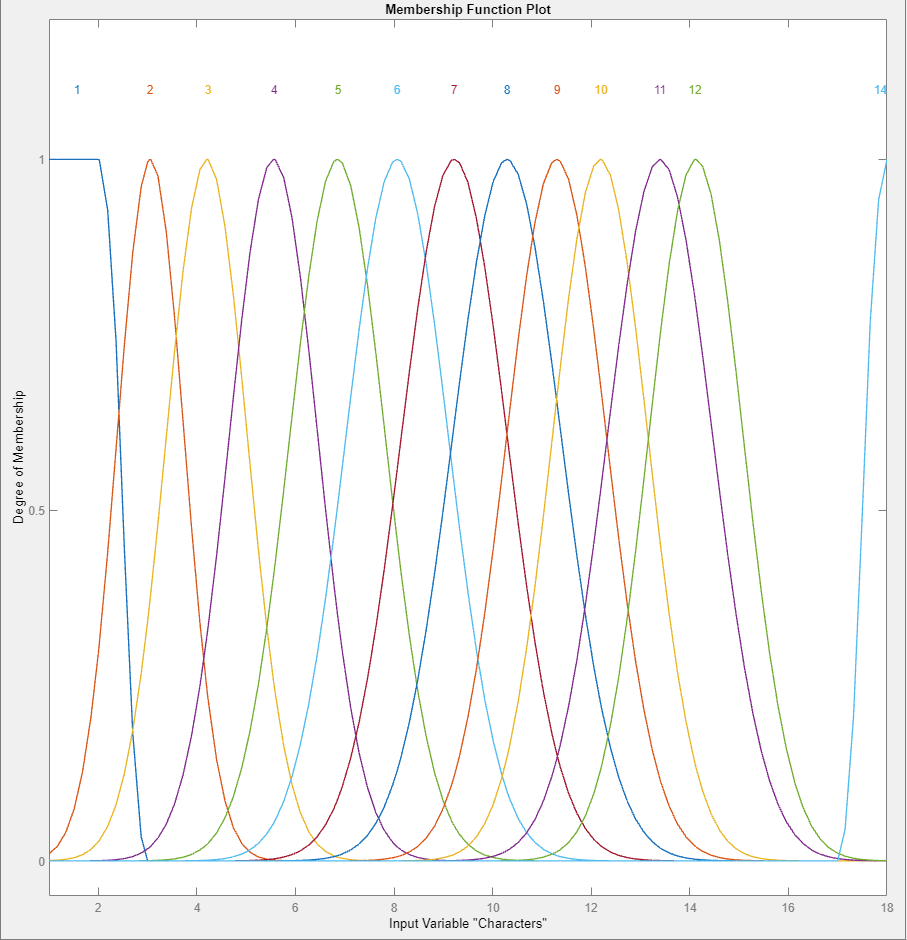}}
\caption{Membership Functions for Grapheme count compared to word length}
\label{CharSets}
\end{figure}

\begin{figure}[htbp]
\centerline{\includegraphics[width=6.5cm]{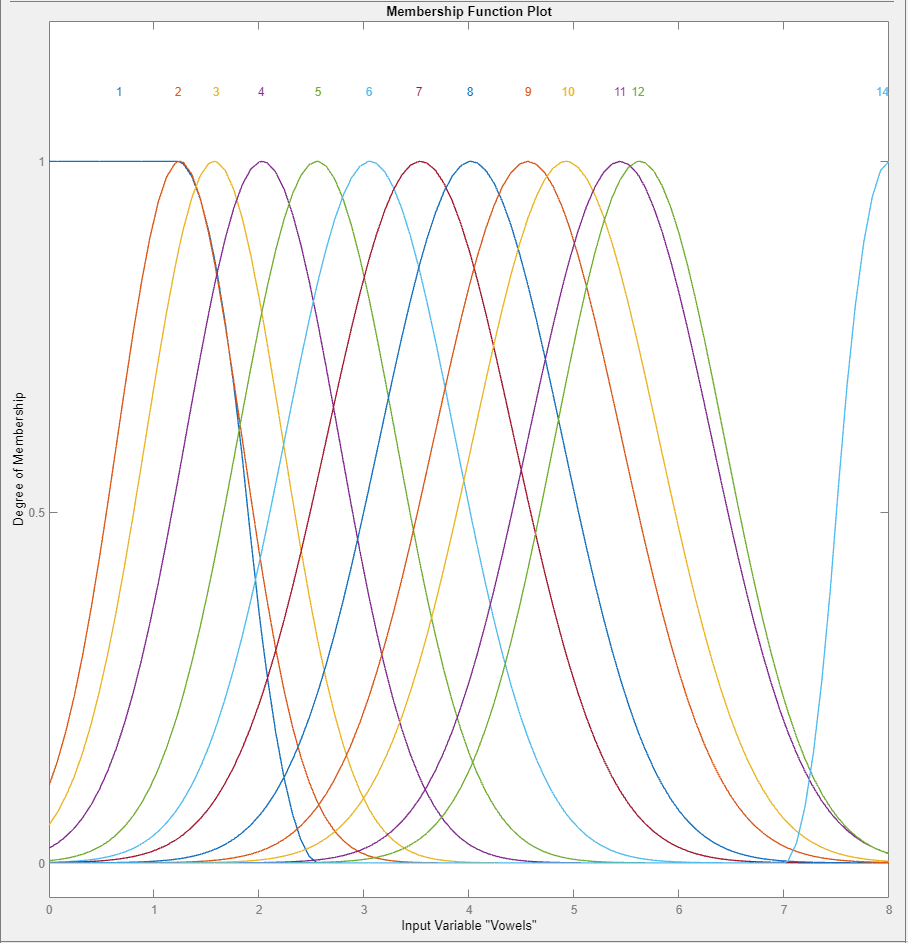}}
\caption{Membership Functions for Grapheme count compared to  a words vowel count}
\label{VowelSets}
\end{figure}

\begin{figure}[htbp]
\centerline{\includegraphics[width=6.5cm]{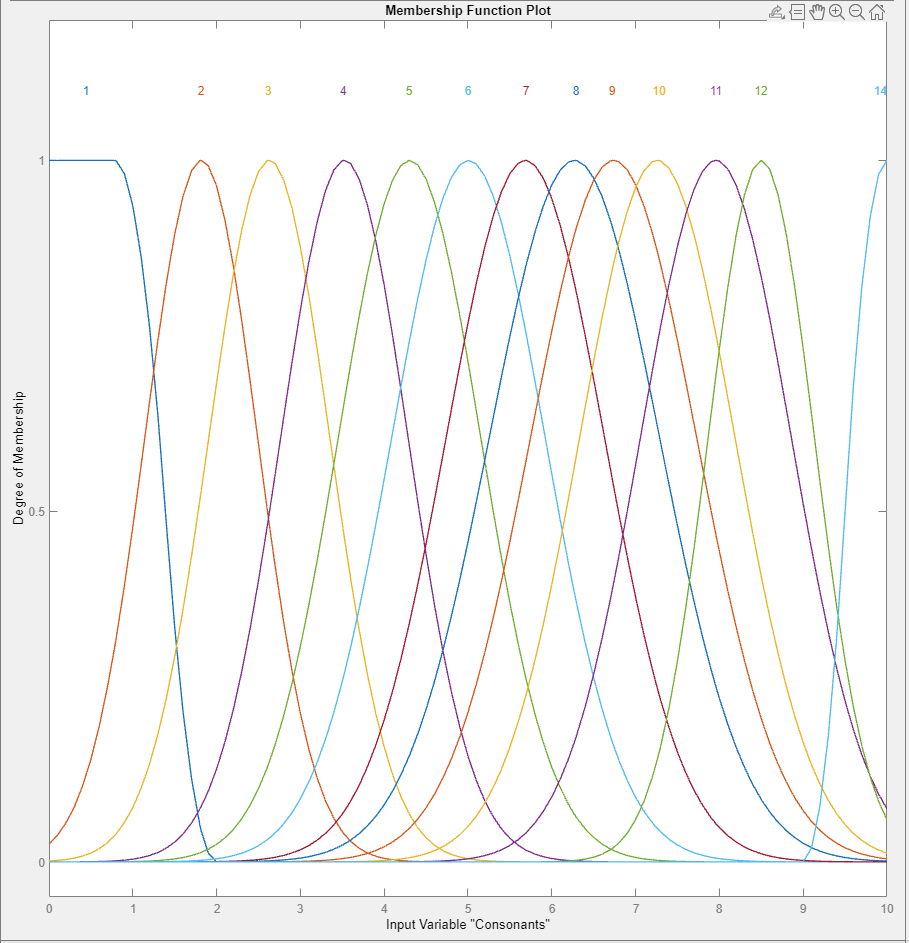}}
\caption{Membership Functions for Grapheme count compared to a words consonant count}
\label{ConsoSets}
\end{figure}

Fuzzy Membership Functions for each feature were created using the mean and standard deviations outlined above in Table \ref{MeanStDevTable}.
These are Gaussian functions that give a membership value in the range [0-1] depending on the input. These curves were used for graphemes 2-12 for all three
features. The sets for 1 \& 14 graphemes were developed as Z and S-shaped curves to catch exceptions at the beginning and end of all three workspaces. This decision was
made based on the characteristics of the corpus and the frequencies shown in Figure \ref{Grapheme Probabilities} which show that all words of length one and a quarter of words length 
two have a single grapheme. Conversely, all words length 18 in the corpus have 14 graphemes. The membership functions for all three features are shown in Figures \ref{CharSets},
\ref{VowelSets}, and \ref{ConsoSets}. While Figure \ref{CharSets} is very distinct, there is a great deal of similarity between Figures \ref{VowelSets} and \ref{ConsoSets}. The main difference being that the Vowel sets have a substantially greater degree of overlap when compared to Consonant sets.

\subsubsection{Output Membership Functions}
\begin{figure}[htbp]
\centerline{\includegraphics[width=7cm]{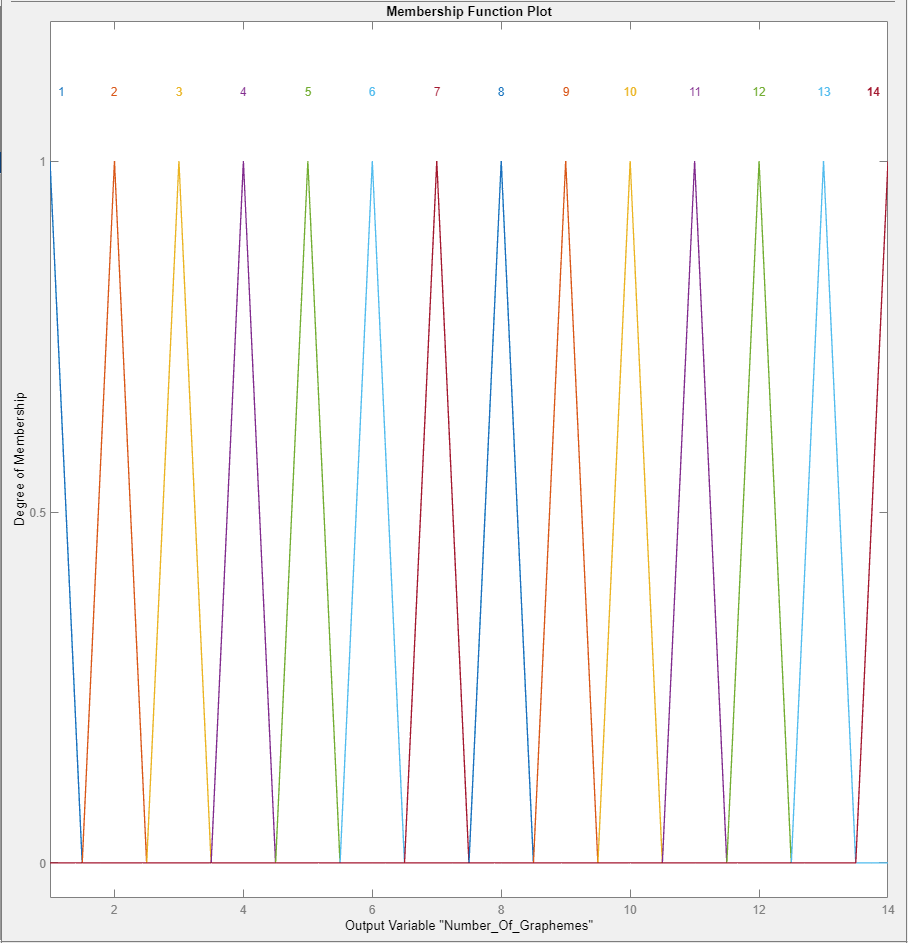}}
\caption{Membership Functions for Grapheme count}
\label{GraphemeSets}
\end{figure}

As each input membership function corresponds to a number of graphemes in a word, the output membership functions are in a workspace between [1-14] representing the number of
graphemes in a word. Because of this, the output membership functions also correspond to the number of graphemes predicted in a word. Unlike the input membership functions, where
the different functions overlap, the output membership functions were defined as discrete triangular functions peaking on the whole number of graphemes that 
function corresponds to, with the tails placed on the edge of where a rounding algorithm would round to that number (0.5 below and 0.499 above). These membership functions are shown 
in Figure \ref{GraphemeSets}.

\subsubsection{Fuzzy Rules}
\label{Rules}
\begin{figure}[htbp]
\centerline{\includegraphics[width=8cm]{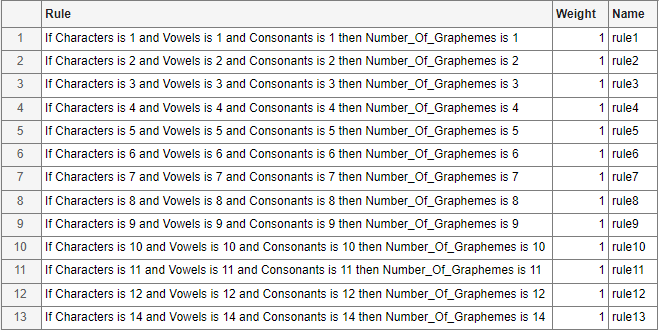}}
\caption{Fuzzy Rules for predicting a words grapheme count}
\label{FuzzyRules}
\end{figure}

Fuzzy rules follow an IF $x$ AND/OR $y$ THEN $z$ format, with $x$ and $y$ being fuzzy input membership functions and $z$ being the fuzzy output membership function \cite{b9}.
In the use case of this project, with the input membership functions directly corresponding to the same feature as the output membership functions, the rules
in this fuzzy inference system follow a simplistic methodology. With 13 input sets for each feature, the 13 rules correspond to output membership 
functions 1-12, and 14. The rules used can be seen in Figure \ref{FuzzyRules}. 

\subsubsection{Grapheme Count Prediction Results} 
The Fuzzy Inference System outputs a number to 2 decimal places in the range [1-14]. This is calculated using the rules outlined above in Section \ref{Rules} followed by a centroid defuzzification method \cite{b14}.
The initial results are listed below in Table \ref{FuzzyResults}.

\begin{table}[]
\centering
\caption{Fuzzy Inference System Results}
\begin{tabular}{|c|c|c|}
\hline
\textbf{Predicted Result} & \textbf{Count}& \textbf{Percentage of Corpus} \\
\hline
Correct Result & 4838 & 50.18\% \\ \hline
One Greater Than Expected       & 2577 & 26.73\% \\ \hline
One Lower Than Expected         & 1600 & 16.6\%  \\ \hline
Prediction Wrong By More Than 1 & 626  & 6.49\%  \\ \hline
\end{tabular}
\label{FuzzyResults}
\end{table}

\subsection{Grapheme Count Mapping}
After the Fuzzy Inference system has predicted the number of graphemes in a specified word, this predicted value is passed into a mapping method. This method
iterates through the list of graphemes, from largest to smallest to find a matching grapheme. This then repeats until the entire word has been split into graphemes.
If at any point the mapping results in a failure to find a valid grapheme, the iteration restarts until a valid combination of graphemes has been selected. 

\section{Results}
To allow for comparison, a second method, IPA mapping, was developed. This method involved using a pronunciation dictionary to return the IPA characters for the word. This then followed a similar iterative system to find a matching IPA Phoneme from the characters, which have all the same problems as grapheme mapping, and from there, use known correspondences to select the valid grapheme.

Using both methods, two different accuracy scores were measured. Firstly, did the method select the correct number of graphemes, secondly, where the correct number of graphemes were selected, were these the correct mapping for the word. The results of these experiments, shown in Table III, show two contrasting conclusions. Firstly, in its current state, IPA mapping results in a greater proportion of words being successfully split into the correct number of graphemes. However, the second conclusion drawn from these results is that the Fuzzy Inference System, followed by mapping graphemes results in a greater proportion
of words being split into the correct, British received pronunciation, mapping of graphemes.

\begin{table}[]
\caption{Results}
\begin{center}
\begin{tabular}{|c|c|c|c|c|}
\hline
\textbf{Method} & \textbf{Training}& \textbf{Training}& \textbf{Validation}& \textbf{Validation} \\
		      & \textbf{Dataset}& \textbf{Accuracy}& \textbf{Dataset}& \textbf{Accuracy} \\
\hline
IPA&Google&69.27\% &EnglishGrammarHere&71.20\%  \\
Mapping&10,000&36.41\%&10,000&38.94\% \\
\hline
Fuzzy&Google&50.18\%&EnglishGrammarHere&55.68\%  \\
Inference&10,000&62.16\%&10,000&59.93\% \\
\hline
\end{tabular}
\label{resultsTable}
\end{center}
\end{table}

\section{Conclusion}
This paper has presented a methods to split words into graphemes using a Fuzzy Inference System This methods can be used for
phonological word structure analysis in NLP and in NLG applications. The experiments undertaken have shown that the general trend of grapheme count can be predicted using a fuzzy inference system. However, the fine tuning is dependant on the specific dialect of the transcriber/reader and therefore a definitive count of N graphemes in a word cannot be classified using a fuzzy inference system and would require manual inference to classify precisely. 

Comparing the two methods, one of the limitations of the IPA mapping method is that it only works with linguistically correct spelling due to relying on the use of a dictionary
for converting a word into IPA. Where words are not in the dictionary, or are spelled incorrectly due to a typo or lack of phonological awareness due to another reason such as Specific Learning Disorders. The fuzzy methods developed counter this issue by not being tied to a dictionary lookup or other limiting resource. Instead they allow for breaking up words into graphemes regardless of 
correctness in spelling.

\end{document}